\documentclass{article}

    \usepackage[nonatbib,final]{tackling_climate_workshop_style}

\usepackage[utf8]{inputenc} 
\usepackage[T1]{fontenc}    
\usepackage{booktabs}       
\usepackage{amsfonts}       

\usepackage{graphicx}
\usepackage{amsmath}
\usepackage[bb=boondox]{mathalfa}
\usepackage{amsthm}
\usepackage{amssymb}

\usepackage{stmaryrd}

\usepackage{amsfonts}
\usepackage{enumerate}
\usepackage{euscript}
\usepackage{mathtools}
\usepackage{nicefrac}
\usepackage[dvipsnames,xcdraw]{xcolor}
\usepackage{multirow}

\usepackage{soul}

\usepackage{tikz}
\definecolor{blue}{HTML}{032F99}
\definecolor{red}{HTML}{AA4A44}
\definecolor{yellow}{HTML}{FDDA0D}

\newtheorem{remark}{Remark}
\newtheorem{observation}{Observation}

\newcommand{\minimize}[1]{\underset{{#1}}{\text{minimize}}}
\newcommand{\maximize}[1]{\underset{{#1}}{\text{maximize}}}

\newcommand{\subjectto}{\text{subject to}}

\newcommand\norm[1]{\lVert#1\rVert}

\title{Price-Aware Deep Learning for Electricity Markets}

\author{%
  Vladimir ~Dvorkin
    \\
  Massachusetts Institute of Technology\\
  Cambridge, MA 02109 \\
  \texttt{dvorkin@mit.edu} \\
  \And
  Ferdinando Fioretto \\
  University of Virginia \\
  Charlottesville, VA 22903 \\
  \texttt{fioretto@virginia.edu} \\
}

\begin{document}

\maketitle

\begin{abstract}
  While deep learning gradually penetrates operational planning of power systems, its inherent prediction errors may significantly affect electricity prices. This paper examines how prediction errors propagate into electricity prices, revealing notable pricing errors and their spatial disparity in congested power systems. To improve fairness, we propose to embed electricity market-clearing optimization as a deep learning layer. Differentiating through this layer allows for balancing between prediction and pricing errors, as oppose to minimizing prediction errors alone. This layer implicitly optimizes fairness and controls the spatial distribution of price errors across the system. We showcase the price-aware deep learning in the nexus of wind power forecasting and short-term electricity market clearing.
\end{abstract}

\section{Introduction}

Addressing the gap between electricity market clearing and power system operations with large shares of renewables, recent literature proposes various deep learning tools for forecasting weather-dependent generation and loads, set-points of conventional generation, and active sets of network constraints, among other applications. Deep learning models learn complex physical processes from an abundance of operational data and instantly map contextual features into informed inputs for market clearing \cite{ferrando2023physics}, and even directly estimate market-clearing outcomes from data \cite{liu2022topology}. However well-informed and tuned these models may be, they are still prone to prediction errors that affect market outcomes for electricity producers and consumers. 

This paper uncovers the propagation of prediction errors of deep learning into locational marginal prices -- the ultimate indicator defining electricity market outcomes across the network -- and demonstrates spatial disparity of these errors across the range of power system benchmark systems. To reduce the errors and improve fairness in electricity market applications, we propose to directly incorporate market-clearing optimization as a deep learning layer to inform predictions on pricing errors. For clarity, we use wind power forecasting as the deep learning task, but other prediction tasks  may benefit from the same approach. Our results show that even small errors cause spatial disparities in congested systems, with usually a small set of buses contributing the most to the overall disparity. While transmission investments would ultimately resolve spatial disparities, informing predictions with market-clearing outcomes already improves fairness. 

\section{Price-Awareness for Fair Deep Learning}

Consider a dataset $\{(\varphi_{i},w_{i})\}_{i=1}^m$ of $m$ operational records, where the $i^{\text{th}}$ record includes a power output $w_{i}$ and the associated feature vector $\varphi_{i}$, collecting weather and turbine data, such as wind speeds and blade pitch angles. The dataset is used to train a deep learning model $\texttt{DeepWP}$ mapping features to wind power outputs, i.e., $\texttt{DeepWP}(\varphi)=\widehat{w}.$ $\texttt{DeepWP}$ is a neural network with a fully connected feedforward architecture, and we train it to minimize the prediction error $\norm{\widehat{w}-w}.$ While standard, this training approach is unaware of the effects of prediction errors on electricity prices.

\subsection{From Prediction to Electricity Price Errors}

To study the impact of prediction errors on electricity prices, we work with the electricity market clearing based on a DC optimal power flow (OPF)  problem \cite{ferrando2023physics,liu2022topology}. Given a wind power forecast $\widehat{w}\in\mathbb{R}^{n}$,  the goal is to identify the conventional generation dispatch $p\in\mathbb{R}^{n}$ which satisfies electric loads $d\in\mathbb{R}_{+}^{n}$ at minimum dispatch costs. Generators produce within technical limits $\underline{p},\overline{p}\in\mathbb{R}_{+}^{n}$ incurring costs, described by a quadratic function with coefficients $c\in\mathbb{R}_{+}^{n}$ and $C\in\mathbb{S}_{+}^{n}$.
The power flows are computed using a matrix $F\in\mathbb{R}^{e\times n}$ of power transfer distribution factors, i.e., the flows are $f=F(p+\widehat{w}-d)\in\mathbb{R}^{e}$ limited by $\overline{f}\in\mathbb{R}_{+}^{e}$. The DC-OPF optimization takes the form:
\begin{subequations}\label{model:DC-OPF}
\begin{align}
    \minimize{\underline{p}\leqslant p\leqslant \overline{p}}\quad& p^{\top}Cp+c^{\top}p\label{DC-OPF:obj}\\
    \subjectto\quad
    &\mathbb{1}^{\top}(p+\widehat{w}-d)=0\;\colon\widehat{\lambda}_{b}, \label{DC-OPF:bal}\\
    &|F(p+\widehat{w}-d)|\leqslant \overline{f}\;\colon\widehat{\lambda}_{\overline{f}},\widehat{\lambda}_{\underline{f}},\label{DC-OPF:flow}
\end{align}
\end{subequations}
which minimizes total costs subject to power balance, and generation and transmission limits. 

The locational marginal prices (LMPs), induced on a particular wind power forecast $\widehat{w},$ are derived from the optimal dual variables associated with power balance and power flow constraints:
\begin{align}
    \pi(\widehat{w}) = \widehat{\lambda}_{b}\cdot\mathbb{1} - F^{\top}(\widehat{\lambda}_{\overline{f}} - \widehat{\lambda}_{\underline{f}}) \in\mathbb{R}^{n},\label{lmp_def}
\end{align}
where the first term is the system-wide price adjusted by the second term due to congestion. 
\begin{remark}[Uniqueness]
    If convex problem \eqref{model:DC-OPF} is feasible for some forecast $\widehat{w}$, then its primal and dual solutions are unique thanks to strictly monotone objective function \eqref{model:DC-OPF} \cite{dvorkin2019electricity}. Hence, electricity price $\pi(\widehat{w})$ is also unique w.r.t. forecast $\widehat{w}$. 
\end{remark}
Hence, the LMP error, defined as
$
    \delta\pi = \pi(\widehat{w}) - \pi(w)
    \in \mathbb{R}^n,
$
is also unique w.r.t. forecast $\widehat{w}$. 

Two key observations arise from analyzing Eq.~\eqref{lmp_def}. The first relates to disparities due to congestion. 
\begin{observation}[Spatial disparity]
   In congested networks, for which $\mathbb{1}^{\top}(\widehat{\lambda}_{\overline{f}}+\widehat{\lambda}_{\underline{f}})>0$, the price error at bus $i$ is proportional to the $i^{\text{th}}$ column of matrix $F$ of power transfer distribution factors. 
\end{observation}
Hence, the same prediction error has a disparate effect on electricity prices, depending on location. This effect under \texttt{DeepWP} predictions is illustrated in Fig. \ref{fig:main} (left plane), where LMP errors vary in the range from $-4.2$ \texttt{\$/MWh} to $+0.9$ \texttt{\$/MWh} on average. 

The second observation identifies a unique system bus, reference bus $r$, with small price errors. 
\begin{observation}[Reference bus]
    Since the $r^{th}$ column of $F$ is all zeros, the price error at the reference bus only includes the error of the system-wide term in \eqref{lmp_def}. 
\end{observation}

The spatial disparities is measured using the notion of $\alpha-$fairness \cite{Fioretto:IJCAIa}: 
\begin{align}
    \alpha=\underset{i\in1,\dots,n}{\text{max}} \norm{
    \mathbb{E}[\norm{\delta\pi_{i}}]-
    \mathbb{E}[\norm{\delta\pi_{r}}]}.\label{alpha-error}
\end{align} 
where the expectation is w.r.t. the dataset distribution. Parameter $\alpha$ is the fairness bound, with smaller values denoting stronger fairness. While the original definition takes the maximum across all pairs of individuals (buses), we use bus $r$ as a reference, as it only includes the error of the system-wide price.

\subsection{Electricity Market Clearing as a Neural Network Layer}

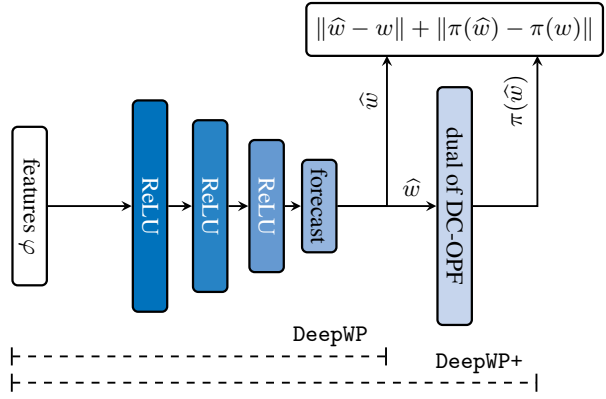
\begin{figure}
\begin{center}
\begin{tikzpicture}[font=\small]
\tikzstyle{input_layer} = [rectangle, rounded corners = 2, minimum width=13, minimum height=60,text centered, draw=black, fill=white!15,line width=0.3mm]
\tikzstyle{1_layer} = [rectangle, rounded corners = 2, minimum width=13, minimum height=80,text centered, draw=black, fill=RoyalBlue,line width=0.3mm]
\tikzstyle{2_layer} = [rectangle, rounded corners = 2, minimum width=13, minimum height=65,text centered, draw=black, fill=RoyalBlue!80,line width=0.3mm]
\tikzstyle{3_layer} = [rectangle, rounded corners = 2, minimum width=13, minimum height=50,text centered, draw=black, fill=RoyalBlue!60,line width=0.3mm]
\tikzstyle{4_layer} = [rectangle, rounded corners = 2, minimum width=13, minimum height=35,text centered, draw=black, fill=RoyalBlue!40,line width=0.3mm]
\tikzstyle{5_layer} = [rectangle, rounded corners = 2, minimum width=13, minimum height=90,text centered, draw=black, fill=RoyalBlue!20,line width=0.3mm]
\tikzstyle{loss_function} = [rectangle, rounded corners = 2, minimum width=90, minimum height=20,text centered, draw=black, fill=white!30,line width=0.3mm]

\node [align=center] at (0.25,3.0) (input) [input_layer] {};
\node [align=center] at (1.85,3.0) (layer_1) [1_layer] {};
\node [align=center] at (2.65,3.0) (layer_2) [2_layer] {};
\node [align=center] at (3.4,3.0) (layer_3) [3_layer] {};
\node [align=center] at (4.1,3.0) (layer_4) [4_layer] {};
\node [align=center] at (5.9,3.0) (opt_layer) [5_layer] {};

\draw[->,>=stealth,line width=0.025cm] (input) -- node[pos=0.5,above] {} (layer_1);
\draw[->,>=stealth,line width=0.025cm] (layer_1) -- (layer_2);
\draw[->,>=stealth,line width=0.025cm] (layer_2) -- (layer_3);
\draw[->,>=stealth,line width=0.025cm] (layer_3) -- (layer_4);
\draw[->,>=stealth,line width=0.025cm] (layer_4) -- node[pos=0.75,above] {$\widehat{w}$} (opt_layer);

\node [align=center,rotate=-90] at (0.25,3.0) {features $\varphi$};
\node [align=center,rotate=-90,white] at (1.85,3.0) {ReLU};
\node [align=center,rotate=-90,white] at (2.65,3.0) {ReLU};
\node [align=center,rotate=-90,white] at (3.4,3.0)  {ReLU};
\node [align=center,rotate=-90] at (4.1,3.0)  {forecast};
\node [align=center,rotate=-90] at (5.9,3.0) {dual of DC-OPF};

\draw[->,>=stealth,line width=0.025cm]  (layer_4) -| (5.,5) node[left,pos=0.85,sloped,above] {$\widehat{w}$};
\draw[->,>=stealth,line width=0.025cm]  (opt_layer) -| (7.,5) node[left,pos=0.85,sloped,above] { $\pi(\widehat{w})$};

\node [align=center] at (5.9,5.35) [loss_function] {$\norm{\widehat{w}-w}+\norm{\pi(\widehat{w})-\pi(w)}$};

\draw[|-|,line width=0.025cm,dashed]  (0,1) -- (5,1) node[pos=0.85,above] {$\texttt{DeepWP}$};
\draw[|-|,line width=0.025cm,dashed]  (0,0.65) -- (7,0.65) node[pos=0.89,above] {$\texttt{DeepWP+}$};

\end{tikzpicture}
\end{center}
\caption{The standard  \texttt{DeepWP} and proposed \texttt{DeepWP+} learning architectures. 
}
\label{fig:nn_architecture}
\end{figure}

To minimize the impact of prediction errors on electricity prices, we develop a new deep learning architecture $\texttt{DeepWP+}$, depicted in Fig. \ref{fig:nn_architecture}. Relative to $\texttt{DeepWP}$, the new architecture includes a market-clearing optimization layer which computes electricity prices in response to predictions and passes them on to the loss function, including both wind power prediction and price errors.

Directly acting on problem \eqref{model:DC-OPF}, however, is computationally challenging: it requires solving the problem  and then differentiating through the large set of Karush–Kuhn–Tucker conditions. We propose to work with the dual optimization instead. For brevity, we rewrite problem \eqref{model:DC-OPF} as
\begin{subequations}\label{model:primal}
\begin{align}
\minimize{p}\quad& p^{\top}Cp + c^{\top}p\\
\subjectto\quad&Ap\geqslant b(\widehat{w})\; \colon \widehat{\lambda},\label{primal:con}
\end{align}
\end{subequations}
where the feasible region is parameterized by forecast $\widehat{w}$. The dual problem takes the form:
\begin{align}\label{model:dual}
\maximize{\lambda\geqslant\mathbb{0}}\quad& q(\widehat{w})^{\top}\lambda - \lambda^{\top}Q\lambda,
\end{align}
where $q(\widehat{w})=AC^{-1}c + b(\widehat{w})$ and $Q=AC^{-1}A^{\top}$. Since the two problems are strictly convex and concave, respectively, the strong duality holds, such that the LMPs can be extracted from the dual problem using expression \eqref{lmp_def}. A virtue of the dual problem \eqref{model:dual} is that it only includes the non-negativity constraints; it is thus simpler to solve and differentiate through than its primal counterpart in \eqref{model:primal}.

The market clearing as an optimization layer is implemented using \texttt{DiffOpt.jl} -- a library for differentiating through the solution of optimization problems in Julia \cite{sharma2022flexible}. 

\section{Numerical Results and Discussion}

For numerical experiments, we use power system benchmark systems from \cite{coffrin2018powermodels}. In each system, we install one wind farm and uniformly scale transmission capacity to provoke congestion, as shown in Table \ref{tab:main}. To ensure fair compassion, we use the same wind power forecasting  data\footnote{https://www.kaggle.com/datasets/theforcecoder/wind-power-forecasting} across all systems. The data includes wind power output as a function of wind speed, wind direction and blade pitch angle features. We independently sample 1,000 scenarios for training and testing.

\begin{table*}[t]
\centering
\setlength\tabcolsep{2pt} 
\caption{Wind power prediction and LMP errors under conventional (\texttt{DeepWP}) and price-aware (\texttt{DeepWP+}) deep learning models}
\label{tab:main}
\resizebox{1\textwidth}{!}{
\begin{tabular}{lrrrrrrrrrrrrrrr}
\toprule
\multirow{3}{*}{case} & \multicolumn{3}{c}{wind power data} & \multicolumn{4}{c}{\texttt{DeepWP}} & \multicolumn{8}{c}{\texttt{DeepWP+}} \\
\cmidrule(lr){2-4}\cmidrule(lr){5-8}\cmidrule(lr){9-16}
 & \multirow{2}{*}{bus}  & \multirow{1}{*}{capacity} & \multirow{1}{*}{$\overline{f}$-scale} & \multirow{1}{*}{RMSE$(\widehat{w})$} & \multirow{1}{*}{RMSE$(\widehat{\pi})$} & \multirow{1}{*}{CVaR$(\widehat{\pi})$} & \multirow{1}{*}{$\alpha-$value} & \multicolumn{2}{c}{RMSE$(\widehat{w})$} & \multicolumn{2}{c}{RMSE$(\widehat{\pi})$} & \multicolumn{2}{c}{CVaR$(\widehat{\pi})$} & \multicolumn{2}{c}{$\alpha-$value} \\
 \cmidrule(lr){3-3}\cmidrule(lr){4-4}\cmidrule(lr){5-5}\cmidrule(lr){6-6}\cmidrule(lr){7-7}\cmidrule(lr){8-8}\cmidrule(lr){9-10}\cmidrule(lr){11-12}\cmidrule(lr){13-14}\cmidrule(lr){15-16}
 &  & \texttt{MW} & [p.u.] & \texttt{MWh} & \texttt{\$/MWh} & \texttt{\$/MWh} & \texttt{\$/MWh} & \texttt{MWh} & \multicolumn{1}{c}{gain}  & \texttt{\$/MWh} & \multicolumn{1}{c}{gain} & \texttt{\$/MWh} & \multicolumn{1}{c}{gain} & \texttt{\$/MWh} & \multicolumn{1}{c}{gain} \\
\midrule
 14\_ieee & 14 & 100 & 1.00  & 0.35 & 0.62 & 1.52 & 0 & 0.35 & $+0.6\%$ & 0.61 &  $\mathbf{-0.6\%}$ & 1.50 & $-0.8\%$ & 0 & ---  \\
 57\_ieee & 38 & 600 & 0.60  & 2.31 & 11.03 & 34.64 & 32.08 & 2.60 & $+11.2\%$ &  10.72 & $\mathbf{-2.9\%}$ & 33.59 & $-3.1\%$ & 30.92 & $-3.8\%$ \\
 24\_ieee & 15 & 1,000 & 0.75  & 4.08 & 8.62 & 37.70 & 27.48 & 4.51 & $+9.6\%$ &  8.33 & $\mathbf{-3.5\%}$ & 36.35 & $-3.7\%$ & 26.26 & $-4.6\%$ \\
 39\_epri & 6 & 1,500 & 0.70  & 5.94 & 11.15 & 31.21 & 17.53 & 6.43 & $+7.6\%$ & 10.19 & $\mathbf{-9.4\%}$ & 28.02 & $-11.4\%$ & 15.84 & $-10.7\%$ \\
 73\_ieee & 41 & 1,000 & 0.80  & 4.02 & 5.12 & 16.21 & 32.83 & 5.51 & $+26.9\%$ & 4.24 & $\mathbf{-20.8\%}$ & 13.41 & $-20.9\%$ & 26.63 & $-23.3\%$ \\
 118\_ieee&  37 & 500 & 0.75 & 2.29 & 3.59 & 11.32 & 17.91 & 2.60 & $+12.1\%$ & 2.88 & $\mathbf{-24.7\%}$ & 9.06 & $-25.0\%$  & 14.09 & $-27.2\%$ \\
\bottomrule
\end{tabular}
}
\end{table*}

\begin{figure*}[t]
    \centering
    \includegraphics[width=1\textwidth]{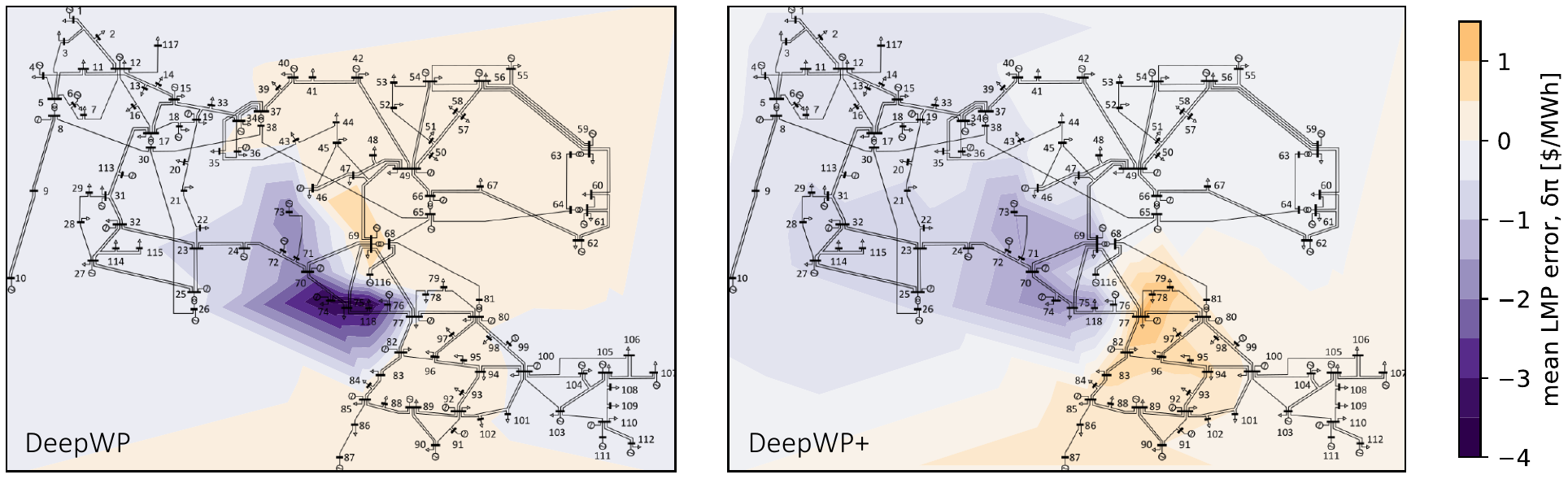}
    \vspace{-0.5cm}
    \caption{Projection of \texttt{DeepWP} and  \texttt{DeepWP+} wind power predictions errors on locational marginal price errors in the IEEE 118-bus system.}
    \label{fig:main}
\end{figure*}

The \texttt{DeepWP} architecture consists of four hidden layers with 30 neurons each, all using ReLU as activation functions. The training consists of three stages, all using ADAM optimizer: pre-training with 500 epochs and learning rate $1e$--$4$, then 1000 epochs with learning rate $5e$--$5$, and final 100 epochs  with learning rate $5e$--$6$. The training of \texttt{DeepWP+} starts from the $501^{\text{th}}$ epoch using the parameters of the pre-trained \texttt{DeepWP} model obtained at the first stage. As neural network parameters are initialized at random, we report the average results across 100 trained models.

We discuss the results using the root mean square error of wind power prediction, denoted $\text{RMSE}(\widehat{w})$, that of prices, denoted $\text{RMSE}(\widehat{\pi})$, the root mean square error across 10\% of the worst-case scenarios, denoted $\text{CVaR}(\widehat{\pi})$, and $\alpha-$fairness bound. 

Table \ref{tab:main} provides a statistical summary for different systems. 
Under prediction errors of the price-agnostic \texttt{DeepWP} model, LMP errors vary between 0.62 and 11.15 \texttt{\$/MWh} 
This range, obtained using the {\it same} wind power dataset, is explained by at least two factors. First, all systems have unique cost structures: the average LMP spans the rage from 42.8 (14\_ieee) to 149.3 (39\_epri) \texttt{\$/MWh}, for the median wind power output. Second, congested networks prone to larger errors. This is particularly evident from the 14\_ieee system, where no wind power output scenario causes congestion. The zero fairness bound $\alpha$ for this system also supports this observation.

The application of the price-aware \texttt{DeepWP+} model, on the other hand, demonstrates the reduction of LMP errors, varying from 0.6 to 24.7\% relative to the  \texttt{DeepWP} model. Even larger reductions are observed across 10\% of the worst-case wind scenarios, measured by  $\text{CVaR}(\widehat{\pi})$. For example, in the 39\_epri case, while the average LMP error is reduced by 0.96 \texttt{\$/MWh}, the worst-case error decreases by 3.19 \texttt{\$/MWh}. Although the focus of \texttt{DeepWP+} is on price errors, it implicitly minimized the disparity of LMP errors, as shown by the reduced fairness bound $\alpha$ across all congested systems. Figure \ref{fig:main} depicts this effect for the most geographically distributed 118\_ieee system, where the fairness bound $\alpha$ is reduced by 27.2\% thanks to price-aware predictions. Observe, while the majority of buses demonstrate near zero price errors, it is a small set of buses contributing to $\alpha-$fairness statistic the most. Importantly, the benefits of \texttt{DeepWP+} do come at the expense of increasing prediction errors. However, all \texttt{DeepWP+} predictions are feasible and yield competitive and fairer -- in the $\alpha-$fairness sense -- electricity prices.

\begin{figure}
    \centering
    \includegraphics[width=0.75\textwidth]{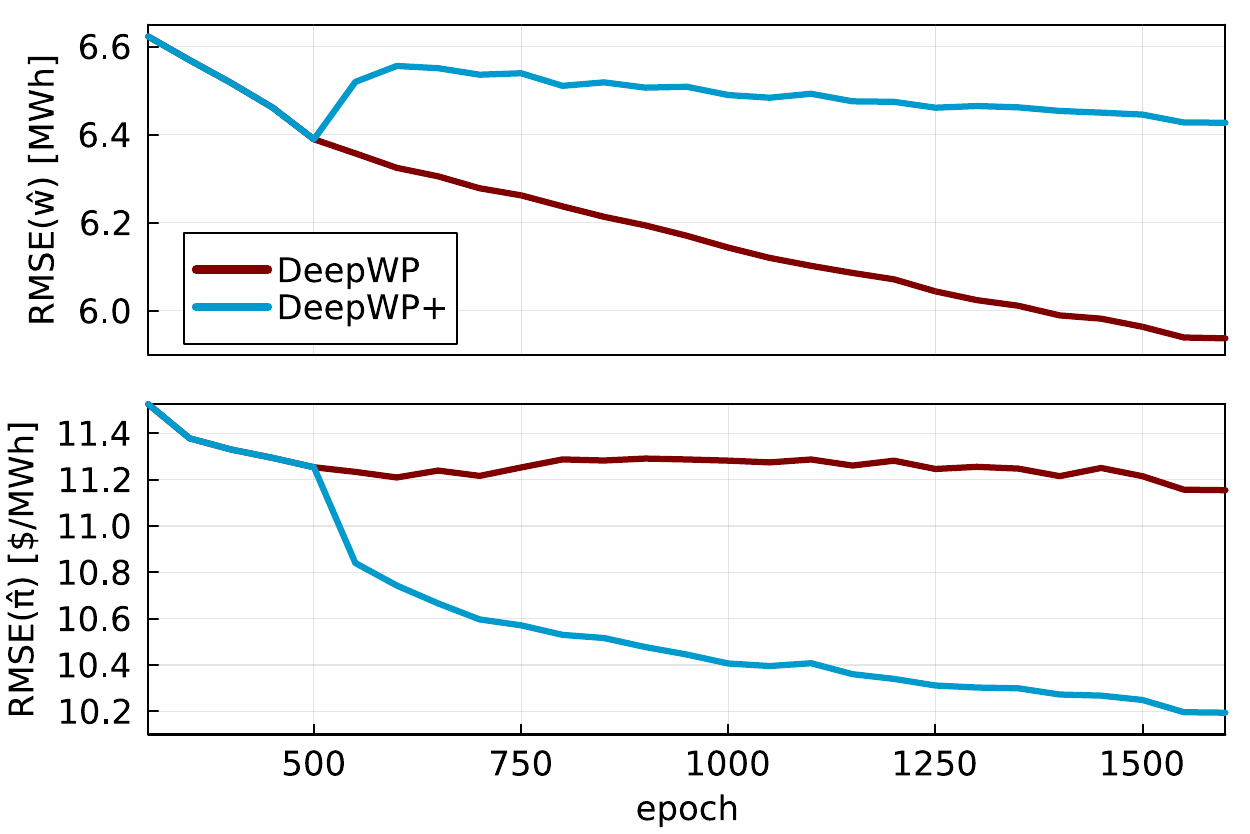}
    \caption{Prediction (top) and locational marginal price (bottom)  errors in the 39-bus EPRI system during the training of \texttt{DeepWP} and \texttt{DeepWP+} models.}
    
    \label{fig:training}
\end{figure}

The price error reduction and fairness improvements are due to the bias which \texttt{DeepWP+} introduces during the training procedure. Figure \ref{fig:training} visualizes how the bias is introduced starting from the $501^{\text{th}}$ epoch, which diverts the training towards the desired result, i.e., to minimal prediction or minimal price error. Finally, we remark that incorporating the electricity market as an optimization layer in deep learning increases the computational burden of the training procedure. Table \ref{tab:CPUtime} reports CPU times per training epoch, which tends to increase in the size of the system. 

\begin{table}
\centering
\setlength\tabcolsep{3 pt}
\caption{Average CPU time per training epoch of \texttt{DeepWP+} architecture}
\label{tab:CPUtime}
\begin{tabular}{lcccccc}
\toprule
case & 14\_ieee & 24\_ieee  & 39\_epri & 57\_ieee & 73\_ieee & 118\_ieee \\
\midrule
time [min.] & 0.06 & 0.26 & 0.22 &  0.48 & 2.03 & 6.28\\
\bottomrule
\end{tabular}
\end{table}

Overall, our results show that embedding market clearing as a deep learning layer informs predictions on market outcomes and improves algorithmic fairness in electricity markets. 

\section*{Acknowledgements}
Vladimir Dvorkin is supported by the Marie Skłodowska-Curie Actions COFUND Postdoctoral Program, Grant Agreement No. 101034297 -- project Learning ORDER.

\bibliography{references.bib}


    

\end{document}